\documentclass[sigconf]{acmart}
\usepackage{subcaption}
\usepackage{listings}
\usepackage{threeparttable}
\graphicspath{ {./figures/} }
\usepackage{fancyvrb} 
\usepackage{booktabs}
\usepackage{multicol}
\usepackage{multirow}

\AtBeginDocument{%
  \providecommand\BibTeX{{%
    \normalfont B\kern-0.5em{\scshape i\kern-0.25em b}\kern-0.8em\TeX}}}

\copyrightyear{2024}
\acmYear{2024}
\setcopyright{rightsretained}
\acmConference[SIGIR '24]{Proceedings of the 47th International ACM SIGIR Conference on Research and Development in Information Retrieval}{July 14--18, 2024}{Washington, DC, USA}
\acmDOI{10.1145/3626772.3661345}
\acmISBN{979-8-4007-0431-4/24/07}

\begin{document}

\title{``Ask Me Anything'': How Comcast Uses LLMs to\\ Assist Agents in Real Time}

\title[How Comcast Uses LLMs to Assist Agents in Real Time]{``Ask Me Anything'': How Comcast Uses LLMs to\\ Assist Agents in Real Time}

\author{Scott Rome}
\email{scott_rome@comcast.com}
\orcid{0000-0003-0270-7192}

\author{Tianwen Chen}
\email{tianwen_chen@comcast.com}
\orcid{0009-0008-0477-7366}
\affiliation{%
  \institution{Comcast AI Technologies}
  \city{Philadelphia}
  \state{Pennsylvania}
  \country{USA}
}

\author{Raphael Tang}
\email{raphael_tang@comcast.com}
\orcid{0009-0007-2873-892X}
\affiliation{%
  \institution{Comcast AI Technologies}
  \city{Philadelphia}
  \state{Pennsylvania}
  \country{USA}
}

\author{Luwei Zhou}
\email{luwei_zhou@comcast.com}
\orcid{0009-0009-1862-3058}

\author{Ferhan Ture}
\email{ferhan_ture@comcast.com}
\orcid{0000-0002-5585-157X}
\affiliation{%
  \institution{Comcast AI Technologies}
  \city{Philadelphia}
  \state{Pennsylvania}
  \country{USA}
}







\renewcommand{\shortauthors}{Rome, et al.}


\begin{abstract}
Customer service is how companies interface with their customers. It can contribute heavily towards the overall customer satisfaction. However, high-quality service can become expensive, creating an incentive to make it as cost efficient as possible and prompting most companies to utilize AI-powered assistants, or "chat bots". On the other hand, human-to-human interaction is still desired by customers, especially when it comes to complex scenarios such as disputes and sensitive topics like bill payment.\footnote{\url{https://bit.ly/3yaNO9t}}


This raises the bar for customer service agents. They need to accurately understand the customer's question or concern, identify a solution that is acceptable yet feasible (and within the company's policy), all while handling multiple conversations at once.

In this work, we introduce ``Ask Me Anything'' (AMA) as an add-on feature to an agent-facing customer service interface. AMA allows agents to ask questions to a large language model (LLM) on demand, as they are handling customer conversations---the LLM provides accurate responses in real-time, reducing the amount of context switching the agent needs. In our internal experiments, we find that agents using AMA versus a traditional search experience spend approximately 10\% fewer seconds per conversation containing a search, translating to millions of dollars of savings annually. Agents that used the AMA feature provided positive feedback nearly 80\% of the time, demonstrating its usefulness as an AI-assisted feature for customer care.

\end{abstract}

\ccsdesc[500]{Information systems~Retrieval models and ranking}
\ccsdesc[500]{Information systems~Language models}

\keywords{rag, llm, customer care, assistive AI, vector db, reranking}

\maketitle

\section{Introduction} 
Comcast, like many other companies, provides customer service through various communication channels. Many self-service solutions are available on the mobile ``Xfinity'' app (e.g., reviewing latest bill) which also has an option to chat with an AI-powered bot named ``Xfinity Assistant''. While these digital automation capabilities have been replacing human customer representatives (also referred to as "agents") for many tasks, there are still many situations that require human-to-human interactions. 

A customer trying to simply look up information about their profile, internet services, or bill, they should be able to do it without an agent's assistance. This also holds true if they are trying to carry out a relatively straightforward task like rescheduling their appointment or make a change to their services. 

Past studies show a human-human interaction is preferred over a human-computer one in certain customer service situations\cite{BehaviorInertia}. For example, agents might outperform bots in situations that require creative problem solving. In other situations, the customer might simply prefer to talk to a agent to benefit from their empathy and emotional intelligence, or to navigate through cultural sensitivities. 


At Comcast, an internal custom tool suite aims to help agents to effectively and efficiently handle such  conversations. However, it still often requires manually looking up information in multiple places, relating it to what the customer is saying, then crafting a relevant response that aligns with the communication guidelines. In this paper, we introduce a new feature to this tool suite called ``Ask Me Anything'' (AMA). It leverages large language models (LLMs) following a retrieval-augmented generation (RAG) approach to generate contextually relevant responses by combining internal knowledge sources, indexing existing knowledge articles efficiently at build time, retrieving relevant chunks of text for a given question at query time, then feeding them to a \emph{Reader} LLM to generate a succinct answer with citations provided as reference. In the next section, we describe the methodology in more detail.
 
\section{Methodology}

Our system follows a typical RAG implementation with modifications to improve performance on proprietary questions. First, the documents are preprocessed to text and chunked, the chunks are embedded then stored with metadata (e.g., associated URL for citations, an identifier, the title, etc.) in a vector database. We describe our specific choices for processing and embeddings in Section \ref{process} and Section \ref{embed} respectively with some experimental justification. Next, we detail how we train and evaluate a reranking model using synthetic data to improve search result relevancy in Section \ref{reranker}. Finally, we discuss how we generate answers followed by how we evaluate the system in Section \ref{gen} and \ref{eval}.

\subsection{Document Preprocessing}\label{process}
We receive documents from various internal clients in different formats. We standardize the documents into plain text and chunk each document into snippets using Deepset.ai's Haystack library \cite{haystack2019}. In order to uniquely reference each chunk of every document after retrieval, we assign an origin identifier to each document and a local identifier to each chunk. Finally, we implement role-based access control on each document, so different users can only view the documents for which they have permission. 

 In \autoref{table:chunker}, we show various chunking parameters for Haystack's preprocessor and their evaluation scores. The metric derivation is explained in Section \ref{eval} (Answer quality assumed the top 3 items were passed to the LLM). We observed a large improvement from setting a higher \verb#max_chars_check#, which we used as a proxy for limiting the size of each snippet given to the LLM. 
 
\begin{table}[H]
    \centering
\caption{Chunking parameters and evaluation of three different settings.}
\begin{tabular}{l c c c}
  \toprule[1pt]
  \textbf{Parameter} & \textbf{A} & \textbf{B} & \textbf{C}\\
  \midrule
  clean\_empty\_lines & true &  \\
  clean\_whitespace & true  & \\
  clean\_header\_footer & true & \\
  split\_by & word & \\
  split\_length & 300 & 100\\
  split\_overlap & 50 & 25 \\
  split\_respect\_sentence\_boundary & true & \\
  max\_chars\_check & 1000 & & 3000  \\ 
    \midrule
\textbf{Metric} &  &   \\
    \midrule
  Answer Quality  & - & -5.7\% & +13.2\% \\
  MRR  & - & -13.3\% & 0.0\% \\
  R@3 & - & -7.9\% & 0.0\% \\
  NDCG & - & -10.0\% & 0.0\% \\
  \bottomrule[1pt] \\
\end{tabular}
\subcaption*{For clarity, only changes from setting $A$ are found in the table. Empty parameter values mean they are same as $A$. The metric values are the relative difference from $A$, i.e., $100\cdot(\mu_B-\mu_A)/\mu_A$ for some metric $\mu$. Metrics are defined in Section \ref{eval}.}
\label{table:chunker}
\end{table}

\subsection{Retrieving Relevant Text Snippets}\label{embed}

To inform the choice of our retriever model, we conducted pilot experiments on a curated evaluation set of fifty question--answer pairs.
We searched the in-production system logs for queries starting with a WH-word (who, what, how, etc.) or ending with a question mark, roughly following the procedure on Bing query logs from \mbox{WikiQA}~\cite{yang2015wikiqa}.
For each question, we then located the relevant passage and answer span in our internal knowledge base used by agents.
Queries without answers were also labeled as such.
Crucially, this process avoids back-formulation~\cite{sakai2004effect}, where queries are manually written by annotators based on known passages rather than crawled from logs, resulting in biased evaluation sets.

We experimented with both dense and sparse retrieval models.
For the sparse model, we used Okapi BM25~\cite{robertson2009probabilistic} with $k_1=1.0$ and $b=0.5$.
For the dense ones, we experimented with four: dense passage retrieval (DPR)~\cite{karpukhin2020dense}, fine-tuned on Natural Questions~\cite{kwiatkowski2019natural}; MPNet-base (v1)~\cite{song2020mpnet}, trained on 160GB of text corpora including Wikipedia, BookCorpus~\cite{zhu2015aligning}, and OpenWebText~\cite{Gokaslan2019OpenWeb}; OpenAI's state-of-the-art \verb#ada-002# embeddings model; and MPNet-base v2, trained further on one billion sentence pairs for better embedding quality.\footnote{Nils Reimers's open-source contribution:\  \url{https://discuss.huggingface.co/t/train-the-best-sentence-embedding-model-ever-with-1b-training-pairs/7354}}
Each was deemed to satisfy our computational and financial constraints at inference time.

In \autoref{tab:retriever-results}, we report the recall@3 (R@3) and the mean reciprocal rank (MRR) of these models on our evaluation set.
The choice of recall@3 (versus recall@5 or 10) is from us feeding the top-three retrieved passages into the LLM.
As a sanity check, we also ran a baseline that randomly drew a passage, which unsurprisingly yielded low scores. 
Mirroring prior work~\cite{yang2019critically}, we found that BM25 remains a strong baseline, outperforming DPR in R@3 and MRR, respectively.
We conjecture that this results from Natural Questions being substantially out of domain from our data.

\begin{table}[H] 
    \centering
    \caption{Results of various retrievers on our pilot evaluation set}
    \label{tab:retriever-results}
    \begin{tabular}{l c c}
        \toprule[1pt]
        \textbf{Method} & \textbf{Recall@3} & \textbf{MRR}\\
        \midrule
        Random & -71.4\% & -83.9\% \\
        BM25 & - & - \\
        \midrule
        DPR (\verb#single-nq#) & -42.8\% & -42.9\% \\
        DPR (\verb#multiset-nq#) & -23.8\% & -29.0\% \\
        Multi-QA MPNet-base & \underline{+33.0\%} & +39.7\% \\
        OpenAI embeddings (\verb#ada-002#) & \underline{+33.0\%} & \underline{+53.9\%} \\
        MPNet-base v2 & \underline{\textbf{+38.1\%}} & \underline{\textbf{+54.9\%}} \\
        \bottomrule[1pt]
    \end{tabular}
    \subcaption*{Statistics presented as relative difference from BM25, i.e., $100\cdot(\mu - \mu_{BM25}) / \mu_{BM25}$ . Underline denotes statistical significance relative to DPR.}
\end{table}

We observe MPNet-base (v1), OpenAI's \verb#ada-002#, and MPNet-base (v2) to perform similarly. Signed-rank tests for R@3 and $t$-tests for MRR also reveal a significant difference ($p<0.05$) from DPR. Due to operational convenience and the high performance of OpenAI's ADA embeddings, we used ADA for the retriever component for the final system.

For our production retrieval step, we embedded both the title of the article and the text of the individual chunk and added them together prior to storage in the vector database. Anecdotally, we found this to yield a more comprehensive retrieval for a variety of queries, especially when chunks were missing some descriptive context of the topic of the article.

\subsection{Reranking Search Results}\label{reranker}

We found that reranking results using models finetuned on synthetic data improved the retrieval step. Our approach was inspired by previous synthetic data generation approaches \cite{bonifacio2022inpars, dai2022promptagator}. First, we used GPT-4 to generate synthetic questions from each snippet in our dataset. We then ran each question through our search system using OpenAI's \verb#text-embeddings-ada-002# \cite{ada} embeddings. Any questions where the original snippet used for question generation did not appear in the top 20 results were discarded. For each synthetic question, we stored the top 20 items retrieved, their relevance as scored by \verb#BGE-reranker-large# \cite{bge_embedding}, and an indicator that the snippet was the source of the question. The final rankings were determined by first placing the source snippet as the "most relevant" result, followed by the snippets in most relevant order as scored by the \verb#BGE-reranker-large# model.

For training, we used RankNet \cite{10.1145/1102351.1102363} to distill these rankings into a finetuned MPNet \cite{song2020mpnet}, in particular \verb#all-mpnet-base-v2# from \verb#sentence-transformer# \cite{reimers-2019-sentence-bert}, which has fewer parameters requiring less computational resources to deploy into production than \verb#BGE-reranker-large#. The final dataset after constructing the necessary pairs for RankNet consisted of over 10 million examples. We set aside 0.5\% of the examples as validation dataset. Our training parameters were listed in Table \ref{table:tr_params}. We used \verb#DistributedDataParallel#
from PyTorch \cite{paszke2017automatic} for distributed training, so the effective batch size is the number of GPUs multiplied by the batch size. We found the "Linear Scaling Rule", where one scales the learning rate when the batch size increases, to not apply to our use case \cite{goyal2018accurate}, but we suspect it is because the original MPNet architecture was trained with a much larger batch size than we used for finetuning.

\begin{table}
\begin{threeparttable}[b]
\centering
\caption{Training hyperparameters.}
\begin{tabular}{l c}
  \toprule[1pt]
  \textbf{Parameter} & \textbf{Specification} \\
  \midrule
  Learning Rate & $5\times 10^{-6}$ \\
  Batch Size & 8 \\
  Number of GPUs\tnote{1} & 10 \\
  Warmup Steps & 4000 \\
  Weight Decay & 0.001 \\
  Epochs & 1\\
  Total Training Steps & 171391 \\
  Learning Rate Scheduler & Warmup-constant \\
  \bottomrule[1pt]
\end{tabular}
\begin{tablenotes}
\item [1] GPU type:\ \texttt{g4dn.xlarge} (Nvidia T4)
\end{tablenotes}
\label{table:tr_params}
\end{threeparttable}
\end{table}

To further evaluate the performance of our reranker model, we randomly sampled 10,000 real questions asked by customer service agents in our production system. For every retrieved document, we followed the approach in \cite{thomas2023large}, which showed that an LLM can accurately predict the relevancy of search results. Specifically, GPT-4 was used to evaluate the overall quality of each document to the question, which combined the scores from how the document matches the intent of the question as well as how trustworthy the document is. The final integer score ranged between 0 and 2, with higher score meaning higher overall quality. Table \ref{table:prod_q_search} compares multiple metrics between ADA vs. reranker. Since the overall score is non-binary, we compute MRR using the rank of first document with a score of 2, and recall@3 examines whether the top 3 documents contain any documents with a score of 2. The results indicate an improvement in retrieval performance with the reranker model.

\begin{table}
\begin{center}
\caption{ADA vs. Reranker Search Results using Production Questions}
\begin{tabular}{l c c}
  \toprule[1pt]
  \textbf{Metric} & \textbf{ADA} & \textbf{Reranker} \\
  \midrule
  Recall@3 & - & +12\% \\
  MRR & - & +15\% \\
  NDCG & - & +4.8\% \\
  \bottomrule[1pt]
\end{tabular}
\subcaption*{For clarity, only changes from setting ADA are found in the table. The metric values are the relative difference from ADA.}
\label{table:prod_q_search}
\end{center}
\end{table}

\subsection{Generating the Answer from Snippets}\label{gen}

In generating the answer, we follow the conventional wisdom approach in the RAG literature. We begin our prompt with a preamble of guidelines for the model, followed by the task description. Due to the length of our snippets of text from the knowledge base, we are unable to provide few-shot examples. We have anecdotally found it better to include more of the text to avoid necessary information being cut off at random. To avoid the "lost in the middle" problem \cite{liu2023lost}, we reverse the order of the Top K results when passed into the LLM, formatted as XML capturing the ID, title and content of the result. We used OpenAI's \verb#gpt-3.5-turbo# for our production Reader component. As a final step in our prompt, we ask the LLM to answer the given question using the search results.


\subsubsection{Citations} 

An important product feature of of the AMA solution is providing references to agents so they can learn more about the answer given. This can be seen in various RAG implementations, such as Microsoft Copilot. In addition, the goal was to build confidence in the system's output and drive adoption internally. Inspired by the Fact-Checking Rail \cite{rebedea2023nemo}, our Citation Rail was accomplished by prompting the LLM to cite its sources in a specific manner (c.f., Figure \ref{gr-prompt}) combined with a post processing step where the citations were removed from the text. If no citations were found, then the system would not return the answer. Practically, there was another benefit from an \textit{observability} perspective:\ through this approach, we identified most "no answer" responses from the LLM, as typically the LLM would response similarly to "I'm sorry. I was unable to find the answer in the documents" without a citation.

\begin{figure}[hbtp]
    \begin{lstlisting}
Please include a single source at the end of your answer, i.e., [Document0] if Document0 is the source. If there is more than one source, use [Document0][Document1] if Document0 and Document1 are the sources.
    \end{lstlisting}
    \caption{An example component of a prompt to encourage citations from the LLM used in the system prompt section.}
    \label{gr-prompt}
\end{figure}

\subsection{Offline Response Evaluation} \label{eval}
To evaluate the system's responses, we follow the LLM-as-a-judge methodology \cite{zhu2023judgelm}, in addition to metrics around retrieval quality typical of a search system. In particular, a random sample of questions from customers were pulled from production traffic. Human annotators then wrote correct answers to each query using internal knowledge bases that are also available to the AMA system. We were able to compare system answers to correct responses given by human annotators using GPT-4 to compute "Answer Quality".

For each question, the annotators also provided a citation from which their answers were based. We used this to calculate "Citation Match Rate":\  the percentage of cases in which the citation from the AMA system matched the ground truth. Given that our retrieval step returns a list, we calculated Recall@K by assuming the annotated citation is the only relevant document. 

Table \ref{table:ama_quality} shows key metrics for the same two approaches as in Table \ref{table:prod_q_search} (\verb #text-embedding-ada-002# for dense retrieval of relevant documents and rerankering the ADA-retrieved documents using our finetuned model). We observe that using reranked documents, LLM is able to achieve a higher answer quality meaning that the answer from a different document ranking is more accurate according to GPT-4. The improvement can also be explained by the increased Citation Match Rate and Recall@3 from the reranked documents directly influencing the LLM's ability to answer accurately.

\begin{table}
\begin{center}
\caption{Response Quality}
\begin{tabular}{l c c}
  \toprule[1pt]
  \textbf{Metric} & \textbf{ADA} & \textbf{Reranker} \\
  \midrule
  Answer Quality & - & +5.9\% \\
  Citation Match Rate & - & +2.5\% \\
  Recall@3 & - & +16.5\% \\
  \bottomrule[1pt]
\end{tabular}
\subcaption*{For clarity, only changes from setting ADA are found in the table. The metric values are the relative difference from ADA.}
\label{table:ama_quality}
\end{center}
\end{table}

\section{Deploying AMA to Customer Service Agents}\label{trial}

Due to business sensitivity purposes, this section will obscure some details related to monetary business metrics. The system was piloted with hundreds of chat agents in late 2023. Over the course of a month-long trial, chat handling time improved 10\% when agents used AMA versus the traditional search option, which required the agent to open a new tool and perform a search. We believe this is a good proxy metric for answer quality because an inaccurate or incomplete response from AMA would require the agent to start over and revert to the traditional option, duplicating work and taking more time overall. Explicit feedback, via a simple thumbs up/thumbs down UI element, was also collected from agents, with nearly an 80\% positive feedback rate (there is no baseline for this rate as such feedback was not requested before the release of this feature). Shortly after the trial period, the system was rolled out to all chat agents (in thousands), with AMA-driven search becoming the preferred way of searching, accounting for two thirds of all typed queries.

\section{Online Reranker Experiment}

Shortly after the trial from Section \ref{trial} concluded, we began an A/B test of the reranker module described in Section \ref{reranker}. The control variant used only the ADA embeddings for vector retrieval with no reranking component, and the treatment utilized the reranker component on the top $20$ results from the ADA-based vector retrieval step. The test ran for three weeks in early 2024. We powered our tests at 80\% and use significance level $\alpha=.01$ for metrics that applied to every interaction and $\alpha=.05$ when metrics considered user feedback, as responses were sparse. Due to the limited pool of agents, our randomization unit, we utilized an agent-day randomization similar to the cookie-day randomization found in other large systems \cite{36500} to increase statistical power. It has been shown in the literature \cite{10.1145/3018661.3018677} that violations of the independent and identically distributed (IID) assumption can lead to underestimation of the variance, but these tests can still be considered trustworthy in practice by using smaller significance thresholds and when observing larger effect sizes. The delta method \cite{10.1145/3219819.3219919} was employed to estimate the variance from question-level metrics. 

We observed a statistically significant increase in two of our metrics:\ namely the "No Answer Rate", which is the number of queries with no answer divided by the total number of queries, and the "Positive Feedback Rate", defined as the number of thumbs up divided by the count of feedback received. Downstream business metrics like average handle time and escalation rate showed no significant difference. However, the improvement in No Answer Rate implies that the system was able to handle more questions than before by providing the relevant documents to the LLM while also increasing the rate of positive feedback.

\begin{table}[H]
\begin{center}
\caption{A/B Test Results}
\begin{tabular}{l c c}
  \toprule[1pt]
  \textbf{Metric} & \textbf{Effect} & \textbf{p-value} \\
  \midrule
  No Answer Rate & -11.9\% & p < .001\\
  Positive Feedback Rate & +8.9\% & p < .05 \\
  \bottomrule[1pt]
\end{tabular}
\subcaption*{ Table contains relative change from control as the effect. Lower is better for No Answer Rate.}
\label{table:ab_test}
\end{center}
\end{table}

\section{Conclusions}

In this paper, we introduced AMA, a large-scale solution to a common business need:\ efficient high-quality customer care. Through the use of third-party LLMs and proven RAG methodology, we were able to build AMA pretty quickly and demonstrate clear value as an assistive feature. We showed improvements to retrieval and answer quality with specific choices for the document preprocessing, the retrieval model and its embeddings, as well as a custom reranker model. As we deploy AMA to thousands of agents with tangible business benefits, we believe that this provides a good  example of how humans and AI can collaborate to better serve customers. 

\bibliographystyle{ACM-Reference-Format}
  \bibliography{bibliography}

\end{document}